\documentclass[letterpaper, 10 pt, conference]{ieeeconf}  % Comment this line out
\IEEEoverridecommandlockouts      
\usepackage{arxiv}
\usepackage{cite}

\title{\LARGE \bf
Active Task-Inference-Guided Deep Inverse Reinforcement Learning
}

\author{Farzan Memarian, Zhe Xu, Bo Wu, Min Wen, Ufuk Topcu \thanks{Farzan Memarian, Zhe Xu and Bo Wu are with the Oden Institute for Computational Engineering and Sciences, University of Texas at Austin, Austin, TX 78712, Ufuk Topcu is with the Department
		of Aerospace Engineering and Engineering Mechanics, and the Oden Institute for Computational Engineering and Sciences, University of Texas at Austin, Austin, TX 78712, e-mails: farzan.memarian@utexas.edu, zhexu@utexas.edu, bwu3@utexas.edu, utopcu@utexas.edu.}
\thanks{Min Wen is with Google LLC. minwen@google.com,}
}

\begin{document}

\maketitle
\thispagestyle{empty}
\pagestyle{empty}

%%%%%%%%%%%%%%%%%%%%%%%%%%%%%%%%%%%%%%%%%%%%%%%%%%%%%%%%%%%%%%%%%%%%%%%%%%%%%%%%
\begin{abstract}
% In inverse reinforcement learning (IRL), given a reward free Markov decision process (MDP) and a set of demonstrations for a task, the objective is to learn the underlying reward function. 

We consider the problem of reward learning for \textit{temporally extended tasks}. 
For reward learning, inverse reinforcement learning (IRL) is a widely used paradigm. Given a Markov decision process (MDP) and a set of demonstrations for a task, IRL learns a reward function that assigns a real-valued reward to each state of the MDP. 
However, for temporally extended tasks, the underlying reward function may not be expressible as a function of individual states of the MDP. 
Instead, the history of visited states may need to be considered to determine the reward at the current state.
To address this issue, we propose an iterative algorithm to learn a reward function for temporally extended tasks.
% At each iteration, the algorithm alternates between a \textit{task inference module} and a \textit{reward learning module}. The task inference module asks a series of queries to be answered by the automaton, and then returns an \textit{automaton} encoding its current hypothesis of the task structure.
At each iteration, the algorithm alternates between two modules,
a \textit{task inference module} that infers the underlying task structure and a \textit{reward learning module} that uses the inferred task structure to learn a reward function. 
The task inference module produces a series of queries, where each query is a sequence of \textit{subgoals}. 
The demonstrator provides a binary response to each query by attempting to execute it in the environment and observing the environment's feedback.
After the queries are answered, the task inference module returns an \textit{automaton} encoding its current hypothesis of the task structure.
The reward learning module augments the state space of the MDP with the states of the automaton.
The module then proceeds to learn a reward function over the augmented state space using a novel \textit{deep maximum entropy IRL} algorithm. 
This iterative process continues until it learns a reward function with satisfactory performance. 
The experiments show that the proposed algorithm significantly outperforms several IRL baselines on temporally extended tasks. 
% At the end of each iteration, the algorithm computes a policy based on the learned reward. 
% This iterative process continues until the computed policy leads to \textit{satisfactory} performance in completing the task. 

% three baselines ({memoryless IRL}, {IRL augmented with information bits } and {memory-based behavioral cloning})

\end{abstract}

% {\color{red}
% \section{To Discuss}
% \begin{itemize}

%     \item I am using task structure instead of temporal structure or memory structure
%     \item We have an algorithm for reward learning module, but no algorithm for task-inference module
% \end{itemize}
% }

\section{Introduction}
In inverse reinforcement learning (IRL)  \cite{ng2000algorithms,abbeel2004apprenticeship,ziebart2008maximum}, given a reward-free Markov decision process (MDP) and a set of demonstrations corresponding to the successful implementation of a task, the goal is to learn the underlying reward function. 
Existing IRL algorithms,
such as \cite{ziebart2008maximum, wulfmeier2015maximum}, learn a reward as a function of the states of the MDP, assuming the reward is independent of the history of states. 
% However, this assumption is not valid for tasks whose objective is to satisfy a co-safe linear temporal logic specification \cite{kloetzer2016multi}; we refer to these tasks as \textit{temporally extended tasks.
However, this assumption may not be valid for \textit{temporally extended tasks}, where we often need to consider the history of states to learn a reward function.

% https://link.springer.com/article/10.1023/A:1018985923441

% whose objective is to complete a set of subtasks in a given temporal order. 
% For such tasks, we often need to consider the history of states to learn a reward function. 

% A simple example of a temporally extended task is the task of reaching a sequence of subgoals in a given order. 

Another challenge for IRL methods is the dependence of the effectiveness of the learned reward function on the demonstrations provided beforehand.
% In this work, we are concerned with reward learning for temporally extended tasks without having access to demonstrations a priori.
To address this issue, \cite{lopes2009active} proposed to actively ask for demonstrations at specific states, and \cite{odom} proposed to actively ask for \textit{advice} in different subsets of the state space. 
% In both cases, the agent targets states or "areas in the state space" that it is uncertain about.
However, for temporally extended tasks, these approaches may not reveal the underlying task structure. Without uncovering the underlying task structure, we may not be able to recover the underlying reward function. %{\color{red} [[[To make this paragraph more clear, I have separated the two works we referenced. One of these works asks for advice. Advice is defined in their work, I can include its definition if you think it's necessary. But it is a 3 line definition. ]]]}

To address the issues discussed above, we develop an iterative algorithm that learns a reward function for temporally extended tasks that are defined over a set of subgoals.
Specifically, the algorithm alternates between two modules: A \textit{task inference module} infers the underlying task structure and a \textit{reward learning module} learns a reward function using the inferred task structure.

The task inference module produces a series of queries to be answered by the demonstrator.
Each query is a sequence of subgoals, and the demonstrator must execute the sequence in the environment and answer the query based on the \textit{binary} feedback from the environment.
For queries for which the environment's feedback is positive, the demonstrator produces demonstrations according to the sequence of subgoals in the query and adds them to the set of demonstrations.
The task inference module then returns a deterministic finite automaton (DFA) encoding the current hypothesis of the task structure.

% At each time step, the state of the product MDP is composed of the current state of the original MDP and the current state of the DFA. 
% The state of the DFA determines the current progress towards task completion and in this sense, the DFA acts as a compact memory structure. 
% The state of the DFA acts as a memory state, i.e., it tracks the progress towards task completion. 
% The module then learns a reward function over the augmented state space by using \textit{deep maximum entropy IRL}. 
The reward learning module augments the state space of the MDP with the states of the DFA.
And then, the module proceeds to learn a reward function over the augmented state space using a novel deep MaxEnt IRL algorithm inspired by \cite{wulfmeier2015maximum}. The IRL algorithm approximates the reward function by a convolutional neural network (CNN).
The reward network takes as input the states of the MDP and the DFA and the current action and outputs a real-valued reward. 
By using a CNN, we avoid the need for manual feature engineering as  CNN can learn to extract reward features. 
The experiments show that the learned reward network generalizes to test environments not seen during training.

In each iteration, after the two modules are called sequentially, the algorithm computes a maximum entropy policy from the learned reward function. 
This iterative process continues until the computed policy leads to a \textit{satisfactory} performance in terms of the probability of completing the task.

To experimentally evaluate the proposed algorithm, we create a \textit{task-oriented navigation} environment. 
We experiment with several navigation tasks and show that the proposed algorithm can successfully infer the underlying task structure and use it to guide IRL. 
We compare the results with two baselines, i.e., a memoryless IRL method, an IRL method augmented with memory bits. 
We show that the proposed method learns policies that outperform the baselines in terms of the probability of completing the navigation tasks. 
We observe the biggest difference in performance between the proposed method and the baselines for navigation tasks with complicated task structures (e.g., the task of reaching a long sequence of subgoals in a given order, where there are repeated subgoals in the sequence).
For such tasks, the policy derived from the proposed method completes the task with a probability that is at least nine times higher than the baselines.

The contributions of the paper are as follows:
\textbf{1)} We develop an algorithm for actively inferring the underlying task structure in the form of a DFA. 
\textbf{2)} We develop a method for inducing the task information encoded as a DFA in deep IRL. 
The DFA tracks the progress in the execution of the task. 
\textbf{3)} 
We propose a novel deep IRL algorithm that learns a reward function over the extended state space. The trained reward network extracts specialized features for each stage of task completion. 

% removes the need for manual feature engineering and improves the generalizability of the method.

\section{Related Work}
% The proposed method has two modules as discussed in previous sections. While both modules are novel, we are not aware of any work combining both of them. We discuss the related work to both of the modules.

%\paragraph{LfD with task inference} 
As part of the pipeline of the proposed algorithm, we learn the task structure and then incorporate it in reward learning.
In prior work such as \cite{niekum2012learning} and \cite{michini2015bayesian} they use assumptions about the task structure to guide policy learning. 
In particular they use Bayesian inference to segment unstructured demonstration trajectories. 
%and use the Beta Process Autoregressive Hidden Markov Model and Dynamic Movement Primitives to learn and generalize a multi-step task for a PR2 robot. 
\cite{shiarlis2018taco} assumed that the expert performs a given sequence of (symbolic) subtasks in each demonstration. 
They solve the problem of temporal alignment for the demonstrations and policy learning for each subtask simultaneously. 
All of these works learn a policy directly, whereas we learn a reward function. 
Moreover, unlike these works, we do not make any assumptions about the task structure prior to training. 
There is prior work attempting to learn the task structure or specialized subtask policies. 
In \cite{henderson2018optiongan}, the authors extend the generative adversarial imitation learning (GAIL) algorithm proposed in \cite{ho2016generative} to learn policy options as well as reward options for complex multi-stage tasks. They claim success in simple and complex continuous control tasks. 
In \cite{kipf2018compile}, they propose a method for learning segments of hierarchically-structured behavior from demonstration data and learn specialized policies or each segment. 
Both previous works assume access to a set of demonstrations a priori, whereas we use an active learning approach.

Our work is closely related to the use of formal methods in RL, such as RL for reward machines \cite{xu2020active,neary2020reward,DBLP:conf/icml/IcarteKVM18} and RL with temporal logic specifications \cite{hasanbeig2018logically,yuan2019modular,hasanbeig2019certified,Fu2014ProbablyAC,tor-etal-aamas18,Min2017,Alshiekh2018SafeRL,LTLAndBeyond,zhe_ijcai2019,tor-etal-aamas18}. For example, in \cite{hasanbeig2019reinforcement}, the authors propose an RL algorithm that learns a policy that maximizes the probability of satisfying a specification expressed as a linear temporal logic formula. 
They construct an automaton based on the linear temporal logic formula and use it to guide the RL agent. The idea of using high-level knowledge such as automaton or temporal logic to guide reinforcement learning \cite{xu2019joint,IcarteNIPS2019} or inverse reinforcement learning has been explored in some previous work. For example, in \cite{xu2019joint}, the authors propose an iterative algorithm that performs joint inference
of reward machines and policies for RL. Our work is different from this line of work as we focus on inverse reinforcement learning rather than reinforcement learning.
In \cite{wen2017learning}, they propose to incorporate side information into the IRL algorithm. Our work is different from this work in that we do not assume any prior knowledge over the task. In addition, they model the reward function as a linear function of known features, whereas we model the reward function by a deep neural network. 

%In their work, the automaton is given beforehand, whereas we learn the automaton as part of an IRL algorithm. 

Our work is also related to hierarchical IRL (HIRL) \cite{krishnan2016hirl}, where they assume each demonstration trajectory corresponds to a set of subtasks. The subtasks are separated by critical \textit{transition states}, i.e., states where transitions between subtasks occur. Once the transition states are recognized, the state space is augmented with features that capture the visitation history of the transition states. HIRL suffers from two limitations compared to the proposed method. First, it assumes that the task can always be decomposed into a sequence of subtasks. This assumption may not hold in general, for example, for tasks that are expressed with more than one sequence of subgoals. Second, their method does not benefit from the use of deep neural networks and automatic feature learning.

\section{Background and preliminaries}
 
The problem of inverse reinforcement learning (IRL) can be described as follows: Given a reward-free MDP $\mathcal{M}$, and a set of demonstration trajectories $D$, learn a reward function $R$ that can optimally interpret the demonstrations in some pre-specified way. 
% Different works on IRL can be distinguished by the following three aspects: First, the reward parameterization; second, the way to generate a policy with a given reward function; third, the interpretation of the demonstrations using the output policy. 

We adopt the following definitions and notations. 
The environment is modeled as a reward-free MDP $\mathcal{M} = \langle S, A,T, \rho, \eta \rangle$ where $S$ is the state space; $A$ is the action space; $T: S \times A \rightarrow \mathcal{D}(S)$ (where $\mathcal{D}(S)$ is the set of all probability distributions over $S$) is the transition function; $\rho \in \mathcal{D}(S)$ is an initial distribution over $S$; and $\eta: S\rightarrow E$ is a labeling function with $E$ as a finite set of subgoals.
Let $D = \{ \tau_1, \ldots, \tau_N\}$ be a set of demonstration trajectories, where $\tau_i = \{(s_{i,0}, a_{i,0}), \ldots, (s_{i, n_i}, a_{i, n_i})\}$ for all $i = 0, \ldots, n_i$. 

We define the task to be learned by a mapping $\mathcal{T}:E^*\rightarrow \{0,1\}$, where $E^*$ is the Kleene star of $E$, and  $\mathcal{T}(\omega)=1$ denotes that a sequence of subgoals $\omega\in E^*$  can complete the task.  
The task structure can be encoded by a deterministic finite automaton (DFA).
A DFA $\mathcal{A}$ is  a tuple $\langle Q_{\mathcal{A}}, \Sigma, \delta, q_0, F \rangle$ where $Q_{\mathcal{A}}$ is a set of states; $\Sigma$ is a set of input symbols (also called the alphabet); $\delta: Q_{\mathcal{A}} \times \Sigma \rightarrow Q_{\mathcal{A}}$ is a deterministic transition function; $q_0 \in Q_{\mathcal{A}}$ is the initial state; $F \subseteq Q_{\mathcal{A}}$ is a set of final states (also called \emph{accepting} states). 
Given a finite sequence of input symbols $w = \sigma_0, \sigma_1, \ldots, \sigma_{k-1}$ in $\Sigma^k$ for some $k \in \mathbb{N}^+$, the DFA $\mathcal{A}$ generates a unique sequence of $k+1$ states $\tau_{\mathcal{A}} = q_0, q_1, \ldots, q_k$ in $Q_{\mathcal{A}}^{k+1}$ such that for each $t = 1, \ldots, k$, $q_t = \delta(q_{t-1}, \sigma_{t-1})$. We denote the last state $q_k$ by taking the sequence $w$ of inputs from $q_0$ as $\underline{\delta}(q_0, w)$. 
$w \in \Sigma^*$ is \emph{accepted} by $\mathcal{A}$ if and only if $\underline{\delta}(q_0, w) \in F$. Let $\mathcal{L}(\mathcal{A}) \subseteq \Sigma^*$ be all the finite sequences of input symbols that are accepted by $\mathcal{A}$, which is also referred to as the \emph{accepted language} of $\mathcal{A}$.

\noindent \textbf{Maximum Entropy IRL.} 
% {\color{red} In this section, we briefly describe the maximum entropy inverse reinforcement learning (MaxEnt IRL) algorithm \cite{ziebart2008maximum}. 
% We show the limitations of this algorithm for learning tasks with complicated task structures, which motivate the proposed algorithm. }
In the original MaxEnt IRL algorithm \cite{ziebart2008maximum}, the reward function is parameterized as a linear combination of a given set of feature functions. 
In other words, given a set of features $\{f_1, \ldots, f_K\}$ where $f_k: S \times A \rightarrow \mathbb{R}$ for $k = 1, \ldots, K$, the reward function $R_\theta$ is parameterized by $\theta = [\theta_1, \ldots, \theta_K]^\intercal$ and $R_\theta(s,a) = \sum_{k=1}^K \theta_k f_k(s,a)$. 
% The basic assumption is that the expected total reward over the distribution of trajectories is the same as the empirical average reward over demonstration trajectories. 
In  MaxEnt IRL, we want to learn a reward function such that the corresponding maximum entropy policy produces the same expected total reward over trajectories as the average reward over trajectories obtained from demonstrations. 
With the principle of maximum entropy, it can be derived that the resulting probability distribution over any dynamically feasible (finite-length) trajectory $\tau = s_0, a_0, \ldots, s_{|\tau|}, a_{|\tau|}$ is proportional to the exponential of the total reward of $\tau$:
\begin{equation}
Pr(\tau | R_\theta) \propto \exp \sum_{(s_i, a_i) \in \tau} R_\theta(s_i, a_i).
\label{eqn:maxentirl_finite}
\end{equation}

While linear parameterization is commonly used in IRL literature \cite{abbeel2004apprenticeship,ratliff2006maximum,neu2007apprenticeship,klein2012inverse}, it suffers from several drawbacks. On the one hand, it requires human knowledge to provide properly designed reward features, which can be labor-intensive; on the other hand, if the given features fail to encode all the essential requirements to generate the demonstrations, there is no way to recover this flaw by learning from demonstrations.
To address these drawbacks, we model the reward function by a deep neural network to automatically construct reward features from expert demonstrations.

\section{Active Task-Inference-Guided Deep Inverse Reinforcement 
Learning}
\label{sec:TODIRL}
In this section, we introduce the active task-inference-guided deep IRL (ATIG-DIRL), which iteratively infers the task structure and incorporates the task structure into reward learning.

\subsection{Overview}
ATIG-DIRL, as illustrated in Algorithm \ref{alg:overall}, consists of a task inference module and a reward learning module. The task inference module utilizes L* learning \cite{angluin1987learning}, an active automaton inference algorithm,  as the template to iteratively infer a DFA from queries and counterexamples. The inferred DFA encodes the high-level task structure to help IRL recover reward functions. Following L* learning, the task inference module generates two kinds of queries with Boolean answers, namely the \emph{membership} query and the \emph{conjecture} query. 
A membership query asks whether a sequence of subgoals can lead to task completion. We defer the details of answering membership queries to Sec.~\ref{subsection:Constructing Hypothesis DFAs}. 
After a number of membership queries are answered, the inference engine outputs a hypothesis DFA and a set of corresponding demonstrations for membership queries that lead to task completion (line~\ref{state:task-inferrence-module} in Alg. \ref{alg:overall}). 
The task inference module then asks a conjecture query about whether the hypothesis DFA can help recover a satisfactory reward function (line~\ref{state:rewardlearningmodule} in Alg.~\ref{alg:irl}). 
The conjecture query is to be answered by the reward learning module, and the details are in Sec.~\ref{sec:reward-learning-module}.
If the answer to the conjecture query is $True$, both the automaton inference process and the reward learning are finished.
Otherwise, the answer is $False$, and there exists a counterexample in the form of a sequence of subgoals to illustrate the difference between the conjectured DFA and the task structure. 
Such a counterexample will trigger a new iteration with the next round of membership queries.

\begin{algorithm}[t]
\caption{Active task-inference-guided deep IRL (ATIG-DIRL)}
\label{alg:overall}
\begin{algorithmic}[1]
\STATE {\bfseries Input:} A reward-free MDP $\mathcal{M} = \langle S, A, T, \rho , \eta \rangle$, stopping threshold $\kappa$ 
\STATE {\bfseries Output:} A hypothesis DFA $\mathcal{A} = \langle Q_{\mathcal{A}}, \Sigma, \delta, q_0, F \rangle$, a reward network $R_\theta: S \times Q_{\mathcal{A}} \rightarrow \mathbb{R}$ and a policy $\pi_\theta: S \times Q_{\mathcal{A}} \rightarrow \mathcal{D}(A)$

\STATE Initialize success ratio $(\beta)$ as 0, counterexample $(CE)$ as null, set of demonstrations $(D)$ as empty set
% \STATE $\beta \leftarrow 0 $; $CE$ $\leftarrow$ None
\WHILE{$\beta$ $ \leq \kappa$}
    \STATE $\mathcal{A}$, $D$ $\gets$ TaskInferenceModule($\mathcal{M}$, $CE$) \label{state:task-inferrence-module}
    \STATE $R_\theta, \pi_\theta$, $\beta$, $CE$ $\gets$ RewardLearningModule($\mathcal{A}$, $\mathcal{M}$, $D$, $\kappa$) \label{state:rewardlearningmodule}
\ENDWHILE
\end{algorithmic}
\end{algorithm}

\begin{algorithm}[t]
\caption{RewardLearningModule}
\label{alg:irl}
\begin{algorithmic}[1]
\STATE {\bfseries Input:} A reward-free MDP $\mathcal{M}$, a hypothesis DFA $\mathcal{A}$, a set of demonstrations $D = \{\tau_1, \ldots, \tau_N \}$, threshold for producing counterexamples $\kappa$
\STATE {\bfseries Output:} Reward network $R_\theta: S \times Q_{\mathcal{A}} \rightarrow \mathbb{R}$, policy $\pi_\theta: S \times Q_{\mathcal{A}} \rightarrow \mathcal{D}(A)$, success ratio ($\beta$),  counterexample ($CE$)
\STATE {\bfseries Hyper-parameters:} number of iterations between updating the success ratio $N$, stopping threshold $\epsilon$
\STATE Initialize the reward network parameter $\theta_0$; Stop $\leftarrow$ False; $t \leftarrow 0$
\WHILE{\textit{Stop = False}}
    \STATE Use the parameter vector $\theta_t$ at the current time step $t$  to compute $Q_{\theta_t}$ and the policy $\pi_{\theta_t}$ via  Eq.~\ref{eqn:soft_Q} and Eq.~\ref{eqn:soft_pi}
    \STATE Compute $\frac{\partial Q_\theta}{\partial \theta} \mid_{\theta = \theta_t}$ and $\frac{\partial \pi_\theta}{\partial \theta} \mid_{\theta = \theta_t}$ via Eq.~\ref{eqn:dQ} and Eq.~\ref{eqn:dpi}
    \STATE Compute $\frac{\partial L_D(\theta_t)}{\partial \theta_t}$ via Eq.~\ref{eqn:dlD} 
%     and then compute $\frac{\partial L(\theta)}{\partial \theta} \mid_{\theta = \theta_t}$.
    \STATE Update $\theta$: $\theta_{t+1} \gets \theta_t + \alpha_t \frac{\partial L_D(\theta)}{\partial \theta} \mid_{\theta = \theta_t}$
    \IF{ $(t \bmod N = 0)$ }
        \STATE Compute the current success ratio $\beta_t$ at time step $t$ using Monte Carlo evaluation \label{state:success-ratio}
        \STATE \textbf{if} ~ $\beta_t - \beta_{t-N} \leq \epsilon$~ \textbf{then}~ \textit{Stop} $\leftarrow$ \textit{True} \label{state:stopping}
    \ENDIF
    \STATE $t\gets t+1$
\ENDWHILE
\IF{ ($\beta_t \leq \kappa$ )}
    \STATE Produce a $CE$ using Monte Carlo simulation \label{state:counterexample}
\ENDIF
\STATE $\beta\gets\beta_t$
\end{algorithmic}
\end{algorithm}

\subsection{Task Inference Module}
\label{subsection:Constructing Hypothesis DFAs}
To generate a hypothesis DFA $\mathcal{A} = \langle Q_{\mathcal{A}}, \Sigma, \delta, q_0, F \rangle$ where $\Sigma = E$,  ATIG-DIRL produces a number of membership queries. 
Each membership query asks whether following a subgoal sequence $\omega\in E^*$ leads to task completion, i.e. $\mathcal{T}(\omega)=1$. 
The task $\mathcal{T}$ is unknown, but given a sequence $\omega\in E^*$, one can observe $\mathcal{T}(\omega)$ from the environment. 
To answer a membership query, we rely on a demonstrator to make a demonstration in the MDP environment and generate a state sequence $\lambda\in S^*$ where $\eta(\lambda)=\omega$.  
Then if the subgoal sequence $\omega$ completes the task, i.e. $\mathcal{T}(\omega)=1$, the answer to this membership is $True$, otherwise the answer is $False$. 
If it is not possible to execute the query, the answer is $False$. 
After answering the membership queries, ATIG-DIRL will generate a hypothesis DFA $\mathcal{A}$ following procedures in \cite{angluin1987learning}.

After obtaining a hypothesis DFA $\mathcal{A}$, ATIG-DIRL asks whether $\mathcal{A}$ is sufficient to recover the reward function (conjecture query). To answer this query, we follow the procedure introduced in Sec.~\ref{sec:reward-learning-module}, which is summarized in Alg.~\ref{alg:irl}.

\noindent \textbf{Demonstration trajectories.}
When the task inference module asks a query $\omega$, if  $\mathcal{T}(\omega)=1$, it means the query encodes a subgoal sequence that leads to task completion. The demonstrator will then produce several demonstrations following the same subgoal sequence and adds them to the set of demonstrations.

\subsection{Reward Learning Module}
\label{sec:reward-learning-module}
This module is concerned with learning a reward function, which is modeled as a deep neural network, using the hypothesis DFA. 
Although previous deep MaxEnt IRL methods \cite{wulfmeier2015maximum,finn2016guided,wulfmeier2016watch,wulfmeier2017large} can learn deep reward networks from demonstrations, they assume the reward function is a function of the current MDP state. 
As a result, the learned policy has to be independent of the history of states, which does not suffice for temporally extended tasks. 
To address this issue, we propose a new maximum entropy deep IRL algorithm (which is inspired by the work in \cite{wulfmeier2015maximum}), as described in Alg.~\ref{alg:irl}.

The key idea is to extend the state space of the MDP using the states of the hypothesis DFA $\mathcal{A}$ and to create a product MDP and then learn a reward function over the extended state space. 
We define the product MDP as follows:

\noindent \textbf{Product MDP.}
Let $\mathcal{M} = \langle S, A,T, \rho, \eta \rangle$ be a reward-free MDP and  $\mathcal{A} = \langle Q_{\mathcal{A}}, \Sigma, \delta, q_0, F \rangle$ be a DFA. The product MDP $\mathcal{M}_{\rm{p}}:=\mathcal{M}\otimes\mathcal{A}=(Z,Z_0,\delta_{\rm{p}},\eta_{\rm{p}},F_{\rm{p}} )$ is a tuple such that, $Z=S\times Q_{\mathcal{A}}$ is a finite set of states. $Z_0$ is the initial set of states where for each $z_0=(s,q)\in Z_0$, $s\in S_0$, $q=\delta(q_0, \eta(s))$. $\delta_{\rm{p}}$ is the transition function of the product MDP defined as
\begin{align*}
\delta_{\rm{p}}((s,q),&  a,(s',q'))= \\
    &\begin{cases}
        T(s,a,s')  ~~\mbox{if}~q'=\delta(q, \eta(s'));\\
        0 ~~~~~~~~~~~~~~~~\mbox{otherwise};
    \end{cases}   
\end{align*}
$\eta_{\rm{p}}((s,q))=\{q\}$ is a labeling function; and $F_{\rm{p}}=S\times F$ is a finite set of accepting states. 
The state space of the product MDP is essentially an extension of the state space of the original MDP.

We apply the proposed IRL algorithm on the product MDP. As a result, the reward depends on both the current state $s$ in $\mathcal{M}$ and the current state $q$ in $\mathcal{A}$, which together form the current state of the product MDP $z\in Z$. DFA states can be considered as different stages in task implementation. Since the DFA state is an input to the reward function, the learned reward, and the corresponding induced policy will be a function of the stage of the task. 

Unlike MaxEnt IRL where the reward function is modeled as a linear combination of pre-specified features, we model the reward function as a neural network. The set of reward parameters are represented by $\theta$, which is the weight vector of the reward network. The objective is to maximize the posterior probability of observing the demonstration trajectories and reward parameters $\theta$ given a reward structure:
\begin{align}
    % \displaystyle
L(\theta) :&= \log Pr(D, \theta | R_\theta)) \nonumber \\
&= \underbrace{\log Pr(D | R_\theta)}_{L_D} + \underbrace{\log P(\theta)}_{L_\theta}
\label{eqn:L}
\end{align} 
$L_D$ is the log-likelihood of the demonstration trajectories in $D$ given the reward function $R_\theta$. $L_\theta$ can be interpreted as either the logarithm of the prior distribution $P(\cdot)$ at $\theta$ or as a differentiable regularization term on $\theta$. In this work, we assume a uniform prior distribution over $\theta$, so what remains is the maximization of $L_D$. 

Since we apply IRL on the state space of the product MDP, the demonstration trajectories are projected onto the state space of the product MDP, i.e. $\tau_i = \{(z_{i,0}, a_{i,0}), \ldots, (z_{i, n_i}, a_{i, n_i})\}$, where $z_{i,j} = (s_{i,j},q_{i,j})$. Let $\pi_\theta$ be the policy corresponding to $R_\theta$, then $L_D$ can be expressed as
% For the infinite-horizon version, $L_D$ can be evaluated explicitly using  Eq.~\ref{eqn:soft_Q} and Eq.~\ref{eqn:soft_pi}:
\begin{equation}
L_D = \sum_{\tau_i \in D} \sum_{j=0}^{n_i - 1} \log \pi_\theta(a_{i,j} | z_{i,j}) + C,
\label{eqn:LD}
\end{equation}
where $C$ is a constant that is dependent on $D$ and the transition dynamics $\delta_p$ of the product MDP.

The computation of $\pi_\theta$ given $R_\theta$ is essentially a maximum entropy reinforcement learning problem \cite{zhou2018infinite}. 
It can be proved \cite{zhou2018infinite} that for any $R_\theta(z,a)$, there exists a unique function $Q_\theta(z,a)$ which is the fixed point of 
\begin{equation}
\begin{aligned}
 Q_\theta(z,a)  &=  R_\theta(z,a) \\ 
 &+ \gamma \sum_{z' \in Z} \delta_p (z' | z , a) \log \sum_{a'} \exp(Q_\theta(z', a')),
\end{aligned}
\label{eqn:soft_Q}
\end{equation}
where $\gamma$ is the discount factor which is a hyper-parameter. Note that $Q_\theta(z,a)$ is an implicit function of $\theta$, as $R_\theta(z,a)$ is parametrized by $\theta$ and $Q_\theta(z,a)$ is derived from $R_\theta(z,a)$ by Eq.~\ref{eqn:soft_Q}. 
The maximum entropy policy $\pi_\theta$, which is alternatively called the \textit{soft Bellman policy} in \cite{zhou2018infinite}, can be derived from $Q_\theta$ as shown in the equation below
\begin{equation}
\begin{aligned}
\pi_\theta(z | a) = \frac{\exp Q_\theta(z,a)}{\sum_{a'} \exp(Q_\theta(z, a'))}.
\end{aligned}
\label{eqn:soft_pi}
\end{equation}
To find the optimal $\theta$, we perform gradient ascent using the gradient of $L_D$ with respect to $\theta$ expressed as
\begin{equation}
\begin{aligned}
\frac{\partial L_D}{\partial \theta} 
=& \frac{\partial}{\partial \theta} \sum_{i=1}^N \sum_{l=0}^{n_i} \log \pi_\theta(z_{i,l}, a_{i,l}) 
 \\=  &\sum_{i=1}^N \sum_{l=0}^{n_i} \Big( \frac{\partial Q_\theta(z_{i,l}, a_{i,l})}{\partial \theta} - \sum_{a'} w_\theta(z_{i,l}, a') \Big).
\end{aligned}
\label{eqn:dlD}
\end{equation}
where $w_\theta(z,a) := \pi_\theta(a | z) \frac{\partial Q_\theta(z,a)}{\partial \theta}$ for any $z\in Z, a \in A$. To compute the right hand side of Eq.\ref{eqn:dlD}, we compute the gradient of $\pi_\theta$ and $Q_\theta$ as
\begin{equation}
\begin{aligned}
&\frac{\partial Q_\theta(z,a)}{\partial \theta} 
= \frac{\partial R_\theta(z,a)}{\partial \theta}  \\ &+  \gamma \sum_{z'} T(z' | z, a) \sum_{a'} \pi_\theta(a' | z') \frac{\partial Q_\theta(z',a')}{\partial \theta}, 
\end{aligned}
\label{eqn:dQ}
\end{equation}

\begin{equation}
\begin{aligned}
\frac{\partial \pi_\theta(a | z)}{\partial \theta} =& z_\theta(z,a) - \pi_\theta(a | z) \sum_{a'} w_\theta(z, a').
\end{aligned}
\label{eqn:dpi}
\end{equation}
Since $\gamma \in (0,1)$, it can be shown that for any $\theta$, there exists a unique solution $\frac{\partial Q_\theta(z,a)}{\partial \theta}$ which is the fixed point to Eq.~\ref{eqn:dQ}. Therefore, there is also a unique solution $\frac{\partial \pi_\theta(a | z)}{\partial \theta}$ to Eq.~\ref{eqn:dpi}. 

Once we have performed a gradient ascent step using Eq.\ref{eqn:dlD} to update the weight vector $\theta$ of the neural network, we have automatically updated the reward network $R_\theta$.

\noindent \textbf{Monte Carlo evaluation.} Evaluating the task performance of a reward network $R_\theta$, amounts to computing the \textit{success ratio} of the maximum entropy policy $\pi_\theta$ for $R_\theta$. 
We use the Monte Carlo approach to empirically compute the success ratio of the policy $\pi_\theta$. 
Concretely, we use $\pi_\theta$ to produce a set of state sequences $\lambda$, convert the state sequences to subgual sequences, i.e, $\eta(\lambda)=\omega$, and then observe $\mathcal{T}(\omega)$. The ratio of sequences with $\mathcal{T}(\omega)=1$ to the total number of sequences yields the success ratio (line~\ref{state:success-ratio} in Alg.~\ref{alg:irl}). 

\noindent \textbf{Stopping criterion.} After every $N$ iterations of gradient ascent ($N$ is hyper-parameter), the module evaluates the success ratio of the computed maximum entropy policy $\pi_\theta$ for $R_\theta$, and stops the iterations once the success ratio stops changing significantly according to a pre-specified threshold $\epsilon$  (line~\ref{state:stopping} in Alg.\ref{alg:irl}).

\noindent \textbf{Counterexamples.} After we finish the iterations of gradient ascent, if the success ratio of the maximum entropy policy $\pi_\theta$ for the obtained reward network $R_\theta$ is smaller than a threshold $\kappa$, then the module produces a counterexample to be used by the task inference module in the next iteration of Alg.~\ref{alg:overall}. 
To produce the counterexample, we apply Monte Carlo simulations and find an $\omega$ such that $\mathcal{T}(\omega) \neq \xi(\omega, \mathcal{A})$ where $\xi(\omega, \mathcal{A})$ is $1$ when $\omega$ is accepted by the DFA $\mathcal{A}$ and is $0$ otherwise (line~\ref{state:counterexample} in Alg.~\ref{alg:overall}).

\section{Experiments}
\label{sec:experiment}
\begin{figure}[tbp]
\centering
\includegraphics[width=0.9\linewidth]{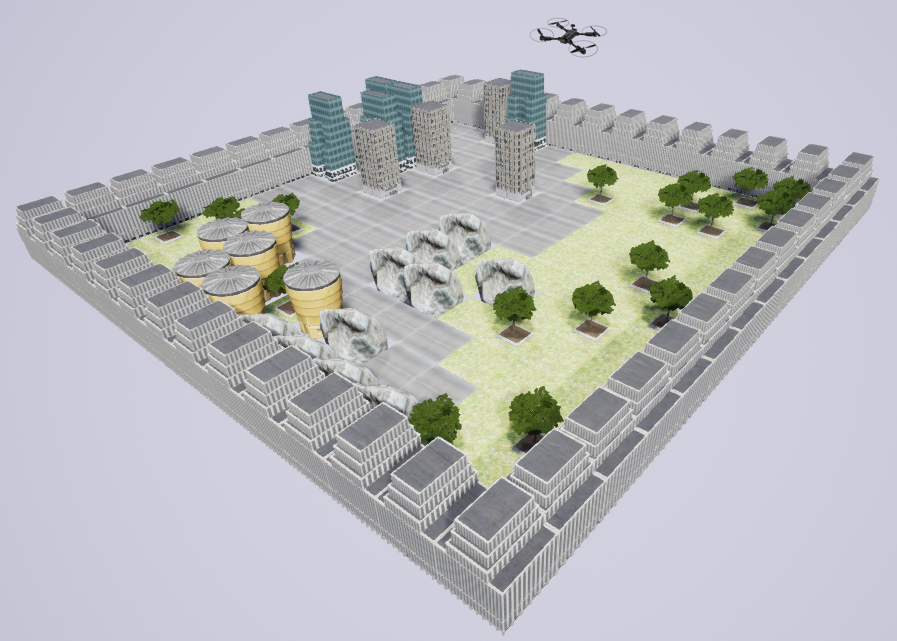}
\caption{Training environment for task-oriented navigation.}
\label{fig:airsim-grid}
\end{figure}

We create the \textit{task-oriented navigation} environment, which is a simulated environment that can model various navigation tasks. 
Different objects are present in the environment, such as \{building, grass, tree, rock, barrel, and tile\}. 
A region is defined as any $3\times 3$ square neighborhood in the environment. 
Each region belongs to one of the types defined in Table.~\ref{tab:region-def}.
Region types are used to define navigation tasks, as we will see later. Visiting a region means visiting the center block of that region. Fig.~\ref{fig:airsim-grid} depicts one instantiation of the environment used as the training environment for our experiments.
  
\noindent \textbf{Agent.} The agent is an aerial vehicle that can navigate in the environment. At each time step, it is located on top of an environment block, and it can choose either of the following actions \{``up", ``down", ``left", ``right"\}, and move one block in that direction. 

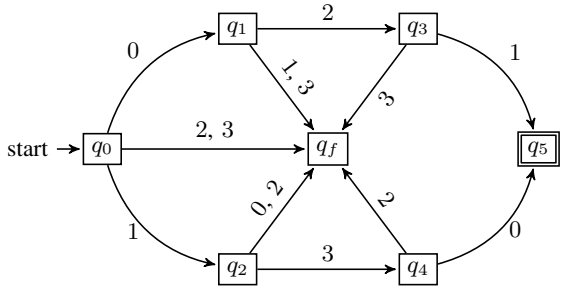
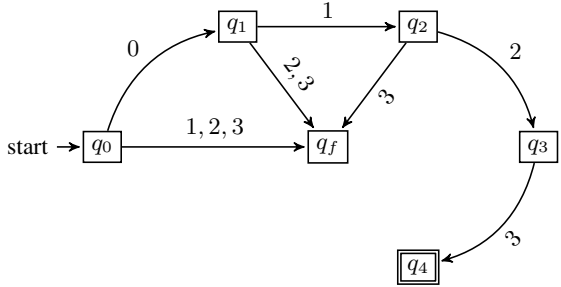
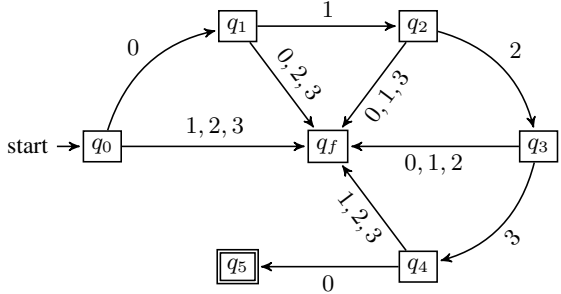
\begin{figure}[h]
\centering
\small
\begin{subfigure}[t]{1\linewidth}
\begin{tikzpicture}[->,>=stealth',shorten >=1pt,auto, semithick, scale=0.8]

 \tikzstyle{state}=[fill=none, draw=black, text=black]
 \tikzstyle{tex}=[fill=none, draw=none, text=black]

 \node[state, initial]           (q0)                   at (0,0)        {$q_0$};
 \node[state]           (q1)     [right of=q0] at (1.,2.0)        {$q_1$};
 \node[state]           (q2)     [right of=q0] at (1,-2.0)         {$q_2$};
 \node[state]           (q3)     [right of=q1] at (4,2.0)        {$q_3$};
 \node[state]           (q4)     [right of=q2] at (4,-2.0)   {$q_4$};
 \node[state, accepting]           (q5)     [right of=q3] at (6,0)   {$q_5$};
 \node[state]           (qf)     [right of=q0] at (2.5,0)   {$q_f$};

 \path    (q0)     edge       [bend left] node {$0$}         (q1)
          (q0)     edge       [bend right] node [left] {$1$}         (q2)
          (q0)     edge       [] node {$2$, $3$}         (qf)
          (q1)     edge       [] node [above, sloped] {$1$, $3$}         (qf)
          (q2)     edge       [] node [above, sloped] {$0$, $2$}         (qf)
          (q3)     edge       [] node [below, sloped] {$3$}         (qf)
          (q4)     edge       [] node [above, sloped] {$2$}         (qf)
          (q1)     edge       [] node {$2$}        (q3)
          (q2)     edge       [] node {$3$}         (q4)
          (q3)     edge       [bend left] node {$1$}         (q5)
          (q4)     edge       [bend right] node [right] {$0$}         (q5);
\end{tikzpicture}
\caption{Underlying DFA corresponding to task $\# 1$.}
\label{subfig:task2DFA}
\end{subfigure}
~
\begin{subfigure}[t]{1\linewidth}
\begin{tikzpicture}[->,>=stealth',shorten >=1pt,auto, semithick, scale=0.8]

 \tikzstyle{state}=[fill=none, draw=black, text=black]
 \tikzstyle{tex}=[fill=none, draw=none, text=black]

 \node[state, initial]           (q0)                   at (0,0)        {$q_0$};
 \node[state]           (q1)     [right of=q0] at (1,2.0)        {$q_1$};
 \node[state]           (q2)     [right of=q0] at (4,2.0)        {$q_2$};
 \node[state]           (q3)     [right of=q0] at (6,0)   {$q_3$};
 \node[state, accepting]           (q4)     [right of=q0] at (4,-2.0)   {$q_4$};
 \node[state]           (qf)     [right of=q0] at (2.5,0)   {$q_f$};

 \path    (q0)     edge       [bend left] node {$0$}         (q1)
          (q0)     edge       [] node {$1,2,3$}         (qf)
          (q1)     edge       [] node {$1$}        (q2)
          (q1)     edge       [] node [above, sloped] {$2,3$}         (qf)

          (q2)     edge       [bend left] node {$2$}         (q3)
          (q2)     edge       node [below, sloped] {$3$}         (qf)

          (q3)     edge       [bend left] node [below, sloped] {$3$}     (q4)  ;         
        %   (q3)     edge       [] node {$0,1,2$}     (qf) 

        %   (q4)     edge       [] node [below, sloped] {$1,2,3$}     (qf);
\end{tikzpicture}
\caption{Underlying DFA corresponding to task $\# 2$.}
\label{subfig:task3DFA}
\end{subfigure}
~
\begin{subfigure}[t]{1\linewidth}
\begin{tikzpicture}[->,>=stealth',shorten >=1pt,auto, semithick, scale=0.8]

 \tikzstyle{state}=[fill=none, draw=black, text=black]
 \tikzstyle{tex}=[fill=none, draw=none, text=black]

 \node[state, initial]           (q0)                   at (0,0)        {$q_0$};
 \node[state]           (q1)     [right of=q0] at (1,2.0)        {$q_1$};
 \node[state]           (q2)     [right of=q0] at (4,2.0)        {$q_2$};
 \node[state]           (q3)     [right of=q0] at (6,0)   {$q_3$};
 \node[state]           (q4)     [right of=q0] at (4,-2.0)   {$q_4$};
 \node[state, accepting]           (q5)     [right of=q0] at (1,-2.0)         {$q_5$};
 \node[state]           (qf)     [right of=q0] at (2.5,0)   {$q_f$};

 \path    (q0)     edge       [bend left] node {$0$}         (q1)
          (q0)     edge       [] node {$1,2,3$}         (qf)
          (q1)     edge       [] node {$1$}        (q2)
          (q1)     edge       [] node [above, sloped] {$0,2,3$}         (qf)

          (q2)     edge       [bend left] node {$2$}         (q3)
          (q2)     edge       node [below, sloped] {$0,1,3$}         (qf)

          (q3)     edge       [bend left] node [below, sloped] {$3$}     (q4)           
          (q3)     edge       [] node {$0,1,2$}     (qf)

          (q4)     edge       [] node [] {$0$}     (q5)           
          (q4)     edge       [] node [below, sloped] {$1,2,3$}     (qf);
\end{tikzpicture}
\caption{Underlying DFA corresponding to task $\# 3$.}
\label{subfig:task1DFA}
\end{subfigure}

\caption{DFAs encoding the underlying task structures. The labels on the edges correspond to the region types. For example $0$ corresponds to $R_0$. 
% Starting from $q_0$, the only way to get to the accepting state $q_5$ is by visiting regions of the types $\{0,1,2,3,0\}$ in this order. 
% Any deviation from this path leads to failure which is captured by $q_f$. 
}
\label{fig:underlying-dfa}
\end{figure}

\noindent \textbf{The navigation task.} %Several navigation tasks can be defined for the agent. 
We experiment with three different navigation tasks. 
The underlying DFA corresponding to these tasks is depicted in Fig.~\ref{fig:underlying-dfa}. 
Among the tasks in Fig.~\ref{fig:underlying-dfa}, task $\#3$ is the most challenging task, and we observe the biggest performance difference between the proposed method and the baselines for this task. 
As such, we explain this task in more detail, and the rest of the tasks follow the same notation and conventions. 
Task $\#3 $ is defined as follows:
\{visit $R_0$, $R_1$, $R_2$, $R_3$, $R_0$ in this order. 
Any other order leads to failure\}.  The states of the underlying DFA for this task are defined in Table \ref{tab:dfa-states}.

With a given task, the goal of ATIG-DIRL is to infer a DFA that is equivalent to the underlying DFA for that task and use it for reward learning. 
We have used the libalf library~\cite{bollig2010libalf} to infer the DFA.

\begin{table}[t]
  \small
  \centering
    \caption{Definition of region types.}
  \begin{tabular}{p{0.3\linewidth}p{0.55\linewidth}}
    \toprule
    % \cmidrule(r)
    Type-0 region~($R_0$) & a region with more than $6$ buildings \\ \midrule
    Type-1 region~($R_1$) & a region with more than $6$ trees\\ \midrule
    Type-2 region~($R_2$) & a region with more than $6$ barrels \\ \midrule
    Type-3 region~($R_3$) & a region with more than $6$ stones \\ \midrule
    Irrelevant region & none of the above  \\
    \bottomrule
  \end{tabular}
  \label{tab:region-def}
\end{table}

\begin{figure}[h]
\centering
\small
\begin{tikzpicture}[->,>=stealth',shorten >=1pt,auto, semithick, scale=0.7]

 \tikzstyle{state}=[fill=none, draw=black, text=black]
 \tikzstyle{tex}=[fill=none, draw=none, text=black]

 \node[state, initial]           (q0)                   at (0,0)        {$q_0$};
 \node[state]           (q1)     [left of=q0] at (-2,2.5)        {$q_1$};
 \node[state]           (q2)     [right of=q0] at (2,2.5)        {$q_2$};
 \node[state]           (q3)     [right of=q0] at (4,0)   {$q_3$};
 \node[state]           (q4)     [right of=q0] at (2,-2.5)   {$q_4$};
 \node[state, accepting]           (q5)     [left of=q0] at (0,-2.5)         {$q_5$};
%  \node[state]           (qf)     [right of=q0] at (2.5,0)   {$q_f$};

 \path    (q0)     edge       [bend left] node {$0$}         (q1)
        %   (q0)     edge       [] node {$1,2,3$}         (qf)
          (q1)     edge       [] node {$1$}        (q2)
          (q1)     edge       [bend left] node {$0,2,3$}        (q0)
        %   (q1)     edge       [] node [above, sloped] {$0,2,3$}         (qf)

          (q2)     edge       [bend left] node {$2$}         (q3)
          (q2)     edge       [] node {$0,1,3$}         (q0)
          
        %   (q2)     edge       node [below, sloped] {$0,1,3$}         (qf)

          (q3)     edge       [bend left] node [below, sloped] {$3$}     (q4)           
          (q3)     edge       [] node {$0,1,2$}         (q0)
        %   (q3)     edge       [] node {$0,1,2$}     (qf) 

          (q4)     edge       [] node [] {$0$}     (q5)  
          (q4)     edge       [] node [] {$1,2,3$}     (q0);  
        %   (q4)     edge       [] node [below, sloped] {$1,2,3$}     (qf)
\end{tikzpicture}

\caption{Intermediate hypothesis DFA for task $\#3$. 
}
\label{fig:intermediate-dfa}
\end{figure}
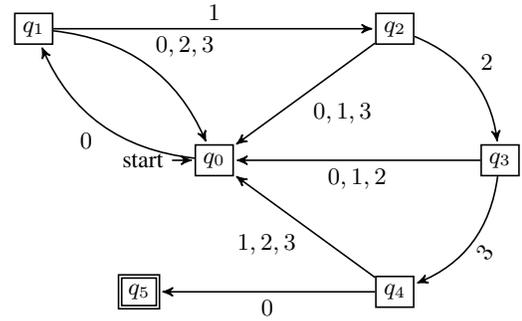

\begin{table}[H]
  \small
  \centering
  \caption{States of the DFA (depicted at Fig.~\ref{subfig:task1DFA}) encoding the underlying task structure. }
  \begin{tabular}{p{0.1\linewidth} p{0.8\linewidth}}
    \toprule
    % \cmidrule(r)
    \textbf{$q_0$} & None of $R_0, R_1, R_2, R_3$ have been visited  \\ \midrule
    \textbf{$q_1$} & Only $R_0$ has been visited  \\ \midrule
    \textbf{$q_2$} & $R_0, R_1$ have been visited in this order \\ \midrule
    \textbf{$q_3$} & $R_0, R_1, R_2$ have been visited in this order \\ \midrule
    \textbf{$q_4$} & $R_0, R_1, R_2, R_3$ have been visited in this order\\ \midrule
    \textbf{$q_5$} & $R_0, R_1, R_2, R_3, R_0$ have been visited in this order  \\ \midrule
    \textbf{$q_f$} & Failure state. Any scenario other than above transits the DFA into $q_f$ (an absorbing state) \\
    \bottomrule
  \end{tabular}

  \label{tab:dfa-states}
\end{table}

\noindent \textbf{Reward network.} 
The input to the convolutional layers of the reward network is the  $7\times7$ neighborhood of the agent in the environment. 
Each block in the neighborhood is represented by a one-hot-encoded vector, specifying the object that occupies that block. 
There are 6 objects, so the input to convolutional layers is a $7\times7\times6$ tensor. 
The network has two convolutional layers, the first layer has $6$ kernels and the second layer has $8$ kernels, all kernels are of size $2\times2$ with a stride of $1$. After the convolutions, there is a flattening layer. 
And this is where the DFA state and the action are provided as input, by appending them to the output of the flattening layer. 
The DFA state and the action are represented by one-hot-encoded vectors of length 10 and 4, respectively. 
Note that the DFAs we learn have at most 7 states, but to make the architecture applicable to bigger DFAs, we use zero padding to make the one-hot-encoded vector be of size 10. 
After appending DFA state and action, the resulting vector passes through two fully connected layers of size $214$ and $50$, and the output layer has one neuron. 
Each hidden layer is followed by a ReLU layer.
% We use gradient ascent for training the reward network, i.e, at each iteration, we use all the demonstrations to calculate $\frac{\partial L_D(\theta)}{\partial \theta}$ according to Eq.~\ref{eqn:dlD}. 

% Each trajectory starts from a random reachable state $(s,q_0) \in S \times Q$ in the PA and stores tuples of the form $(s_t, q_t, a_t)$ for each time step $t$. Concretely, each demonstration trajectory $\tau_i$ is stored as: $\tau_i = \{ (s_t, q_t, a_t) \}_{i=0}^{T_i - 1} $, where $T_i$ is the length of trajectory $\tau_i$. 

\noindent \textbf{Baselines.} We implemented two baseline methods to compare with ATIG-DIRL: \textit{memoryless IRL} and \textit{IRL augmented with information bits (IRL-IB)}. 
The memoryless IRL is a basic deep MaxEnt IRL method \cite{wulfmeier2015maximum} that learns a reward function as a function of the states of the MDP. 
The IRL-IB method is a deep MaxEnt IRL method that uses a fixed size memory to augment the state space of the MDP. In our experiment, since there are 4 region types, the memory is a vector of size 4, where each element corresponds to one of the region types. If a region type has been visited, the value of the corresponding element is $1$, and $0$ otherwise. 
% The memory-based BC method operates on the same extended state space as ATIG-DIRL, and uses the behavioral cloning \cite{pomerleau1991efficient} method to learn a policy that mimics the demonstrations. 
% The memory-based BC method is a simplified version of the method introduced in \cite{shiarlis2018taco} in which the task label is provided for each individual state in the demonstrations, and the agent does not need to learn the alignment between the demonstrations and \textit{task sketches}.

% \subsubsection*{ATIG-DIRL vs baselines} 
\noindent \textbf{Evaluation criterion.}
The primary criterion we use in evaluating the performance of a trained model is the task success ratio. 
All methods are trained using the same training environment (Fig.~\ref{fig:airsim-grid}). 

\noindent \textbf{Training performance.}
We compares the training performance of ATIG-DIRL algorithm to the IRL baselines for the tasks described by the DFAs in Fig.\ref{fig:underlying-dfa}.

The ATIG-DIRL was able to infer a DFA equivalent to the underlying DFA (Fig.~\ref{fig:underlying-dfa}) for all three tasks in at most three iterations of the main algorithm (Alg.~\ref{alg:overall}). 
We have shown an intermediate hypothesis DFA inferred by the algorithm for task $\#3$ in Fig.~\ref{fig:intermediate-dfa}. This DFA recognizes the correct sequence for completing the task, i.e., $R_0, R_1, R_2, R_3, R_0$, but it fails to capture the fact that any deviation from this path results in failure and the task cannot be completed anymore. 
For example, this DFA hypothesizes that the sequence $R_0, R_2, R_0, R_1, R_2, R_3, R_0$ completes the task, whereas by looking at the underlying DFA (Fig.\ref{subfig:task3DFA}), we realize that this sequence leads to failure. 
The reported results for ATIG-DIRL correspond to the last call to Alg.~\ref{alg:irl} where the final hypothesized DFA is used in the IRL loop. 

\begin{figure}[]
\centering
\begin{subfigure}[t]{0.4\textwidth}
\includegraphics[width=\textwidth]{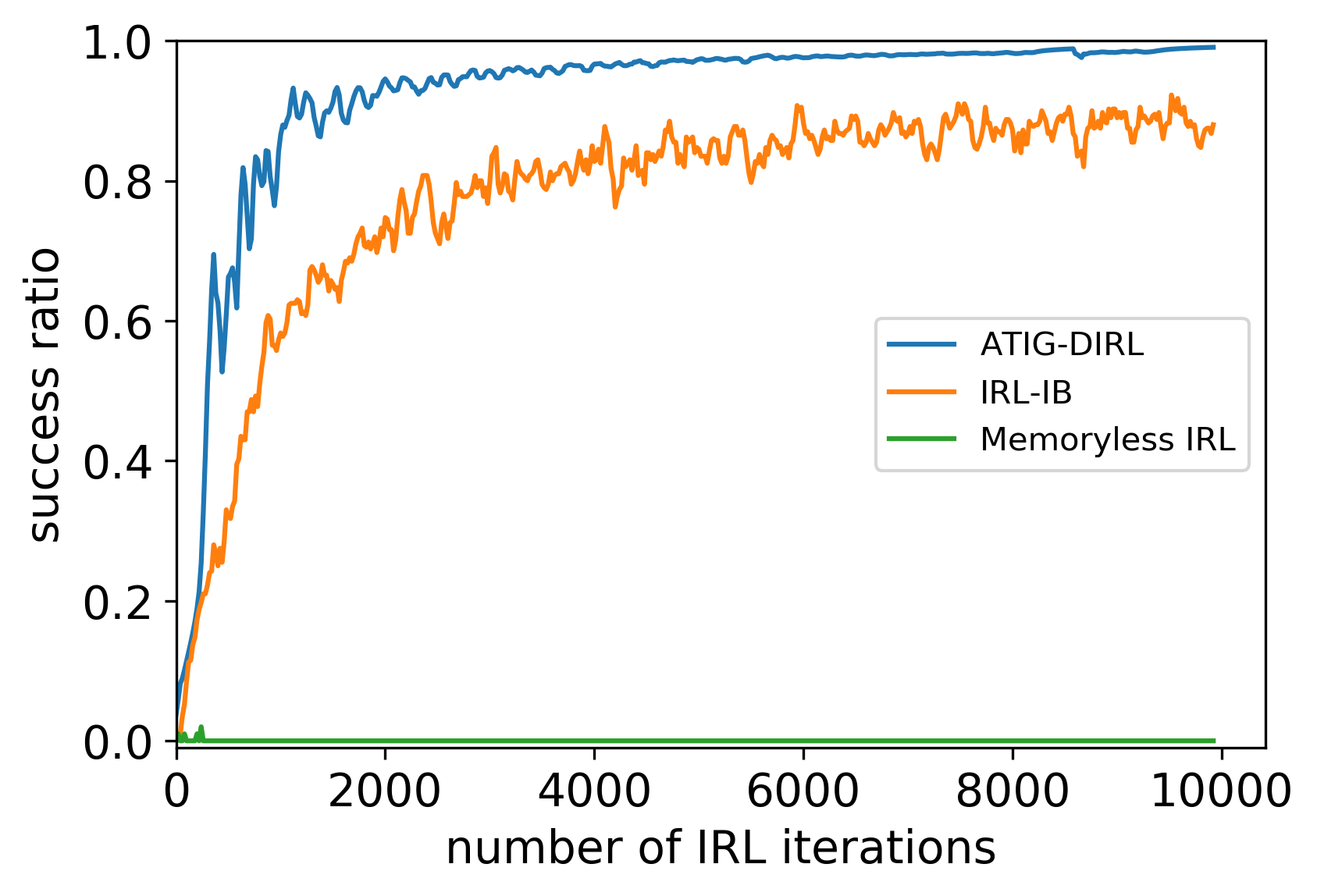}
\caption{Training performance of the baselines for task $\#1$.}
\label{fig:train-task-1}
\end{subfigure}
~~
\begin{subfigure}[t]{0.4\textwidth}
\includegraphics[width=\textwidth]{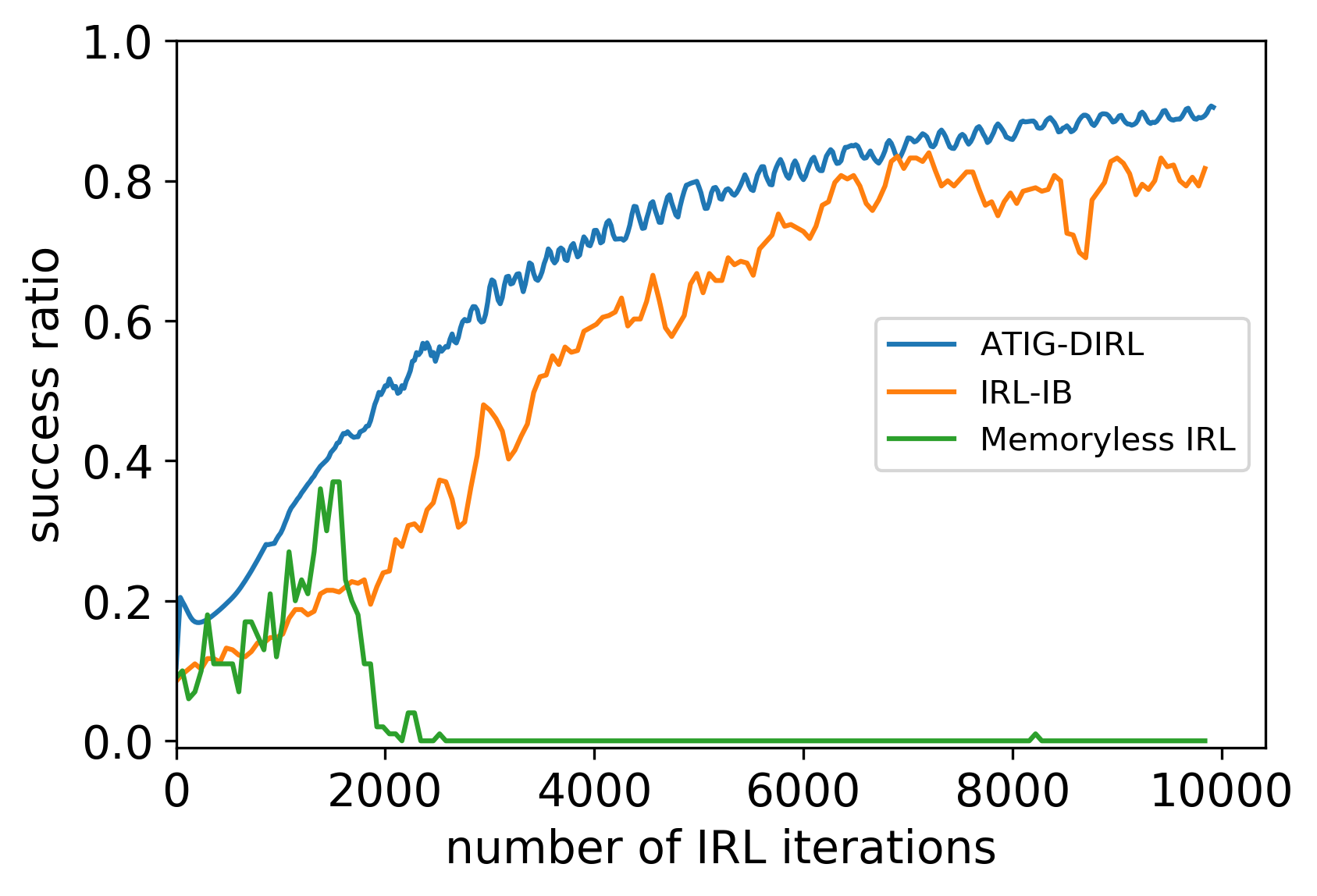}
\caption{Training performance of the baselines for task $\#2$.}
\label{fig:train-task-2}
\end{subfigure}
~~
\begin{subfigure}[t]{0.4\textwidth}
\includegraphics[width=\textwidth]{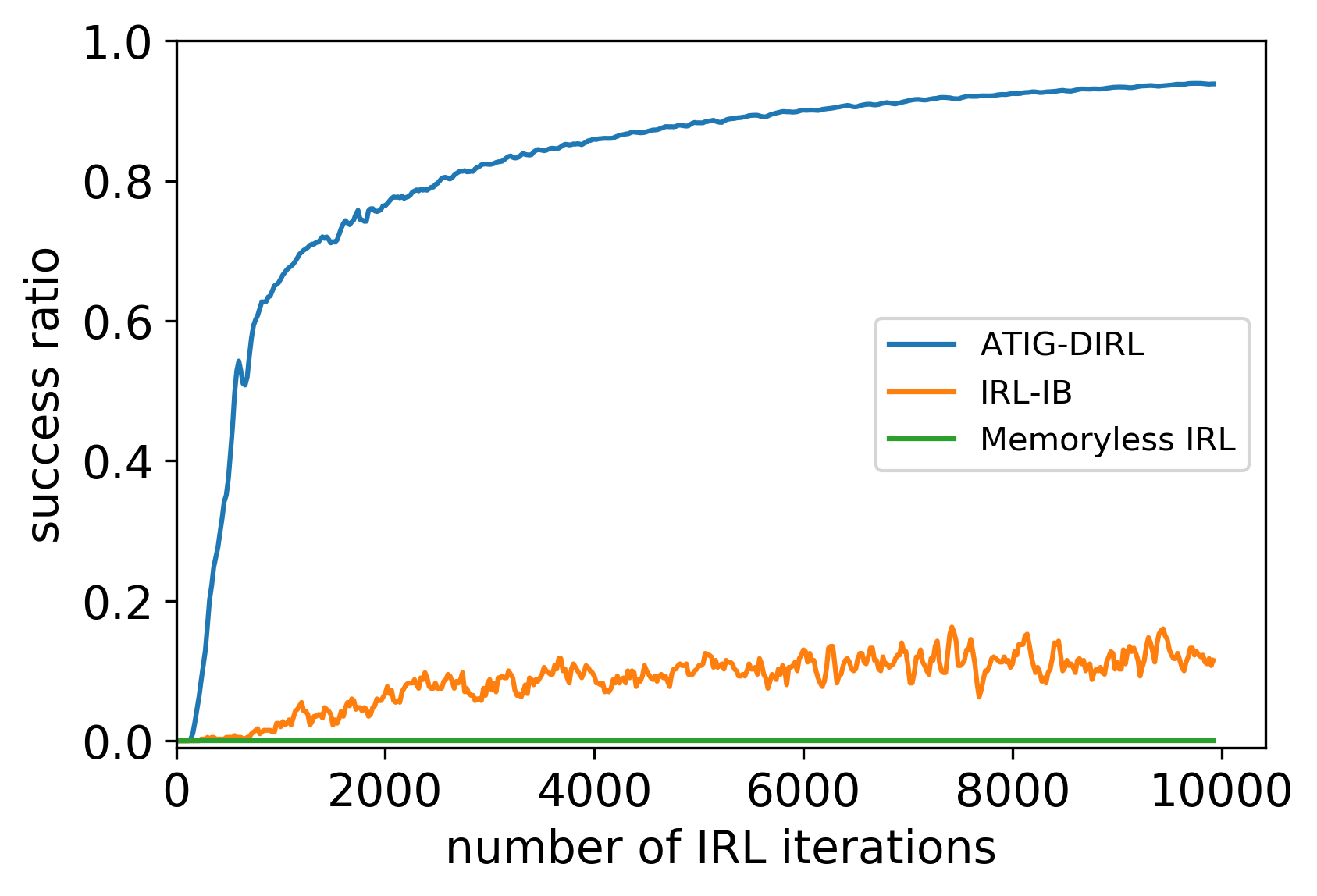}
\caption{Training performance of the baselines for task $\#3$.}
\label{fig:train-task-3}
\end{subfigure}
\caption{Performances of ATIG-DIRL versus the IRL baselines at training time.}
\label{fig:training-irl-baselines}
\end{figure}

Fig.~\ref{fig:training-irl-baselines} shows that as tasks become more challenging from task $\#1$ to task $\#3$, the difference between the performance of ATIG-DIRL and the baselines becomes more pronounced. 
Tasks become more challenging if there are more ways to fail, fewer ways to complete the task, and if the sequences that lead to task completion are longer.
Another factor that makes tasks more complicated is the existence of repeated subgoals in the sequences that complete the task. 
Task $\#3$ is particularly challenging due to all the factors discussed above, particularly the existence of a repeated subgoal ($R_0$) in the sequence that completes the task.  

IRL-IB is able to learn a reward function with comparable performance to the proposed method for tasks $\#1$ and $\#2$, but it fails to do so for task $\#3$. 
The reason is that the memory structure that IRL-IB uses is unable to capture sequences with repeated subgoals.
The memoryless IRL method performs poorly as expected because, without a memory to keep track of task progress, it is not possible to uncover the underlying reward function for the temporally extended tasks considered here. 
% The BC baseline operates on the same augmented state space as our method. 
% As such, it learns a different policy for each state of the DFA. 
% This baseline learns policies with success ratios comparable to the proposed method in the training environment. However, the main drawback of this baseline is in generalizing to test environments. 

\noindent \textbf{Test performance.} 
To compare the generalization capability of the three methods, we have tested them on 10 different randomly generated environments. 
The ATIG-DIRL method outperforms all the baselines on test cases (Tables \ref{tab:todirl-test-task1}, \ref{tab:todirl-test-task2}, \ref{tab:todirl-test-task3}). 
The memoryless IRL method has a poor performance across all tasks as expected because it did not perform well even in the training environment. 
The IRL-IB method, however, performs well for tasks $\#1$ and $\#2$, but fails at generalization for task $\#3$ which is the most challenging task. The main reason for its failure is the existence of a repeated subgoal in the subgoal sequence that completes the task. 
% The BC method performs poorly at test time although it operates on the extended state space similar to ATIG-DIRL. 
% The reason for the difference in generalizability between these two methods is as follows: the ATIG-DIRL algorithm learns a reward function but the BC method directly learns a policy. 
% The reward function is a more compact representation of the desired behavior and hence generalizes to new environments significantly better than a policy. 

\begin{table}[t]
\caption{Mean and standard deviation of success ratio on 10 different test environments for different tasks.}
\begin{subtable}{1\linewidth}
  \centering
  \small
  \caption{Task $\#1$.}
  \begin{tabular}{p{0.4\linewidth}|p{0.12\linewidth} |p{0.12\linewidth}}
    \toprule
    % \cmidrule(r)
    \textbf{Model} & \textbf{Mean } & \textbf{STD } \\ \thickhline
    ATIG-DIRL & $0.95$ & $0.02$ \\ \hline
    IRL-IB & $0.93$ & $0.02$ \\ \hline
    % Memory-based BC & $0.12$ & $0.14$ \\ \hline
    Memoryless IRL & $0.002$ & $0.004$ \\ \hline 
    % \bottomrule
  \end{tabular}
  \label{tab:todirl-test-task1}
\end{subtable}

\medskip

\begin{subtable}{1\linewidth}
  \centering
  \small
  \caption{Task $\#2$.}
  \begin{tabular}{p{0.4\linewidth}|p{0.12\linewidth} |p{0.12\linewidth}}
    \toprule
    % \cmidrule(r)
    % \textbf{Model} & \textbf{Mean } & \textbf{STD } \\ \thickhline
    ATIG-DIRL & $0.95$ & $0.03$ \\ \hline
    IRL-IB & $0.89$ & $0.04$ \\ \hline
    % Memory-based BC & $0.13$ & $0.16$ \\ \hline
    Memoryless IRL & $0.02$ & $0.03$ \\ \hline 
    % \bottomrule
  \end{tabular}
  \label{tab:todirl-test-task2}
\end{subtable}

\medskip

\begin{subtable}{1\linewidth}
  \centering
  \small
  \caption{Task $\#3$.}
  \begin{tabular}{p{0.4\linewidth}|p{0.12\linewidth} |p{0.12\linewidth}}
    \toprule
    % \cmidrule(r)
    % \textbf{Model} & \textbf{Mean } & \textbf{STD } \\ \thickhline
    ATIG-DIRL & $0.96$ & $0.02$ \\ \hline
    IRL-IB & $0.11$ & $0.06$ \\ \hline
    % Memory-based BC & $0.11$ & $0.12$ \\ \hline
    Memoryless IRL & $0.003$ & $0.006$ \\ \hline
    % \bottomrule
  \end{tabular}
  \label{tab:todirl-test-task3}
\end{subtable}
\end{table}

\section{Conclusion and Future Work}
We have proposed a new IRL algorithm, active task-inference-guided deep IRL (ATIG-DIRL). 
The algorithm learns the task structure in the form of a deterministic finite automaton (DFA) and uses the DFA to extend the state space of the original MDP and learns a reward function over the extended state space. 
% The proposed algorithm learns a reward function over the extended state space obtained by composing the state spaces of the MDP and the inferred DFA. 
By representing the reward functions as a convolutional neural network, the algorithm can automatically extract reward features that are essential for task implementation. 
We experiment with various navigation tasks and show that the learned reward function can be used to generate policies that outperform the baselines in both training and test environments. 
% In the most challenging task we tried, the proposed method is the only method that can learn a reward function with satisfactory performance as compared with the baselines. 
% The baseline methods IRL with information bits and memoryless IRL  have poor generalization performance. 
For future work, we will apply our algorithm to more complex environments with continuous state spaces.

\bibliographystyle{IEEEtran}
\bibliography{main}

\end{document}